\documentclass[letterpaper, 10 pt, conference]{ieeeconf}  

\IEEEoverridecommandlockouts                              

\usepackage{graphics} 
\usepackage{epsfig} 
\usepackage{amsmath} 
\usepackage{amssymb}  
\usepackage{mathtools}
\usepackage{colortbl}
\usepackage{slashbox}

\newcommand\kevinupdate[1]{\textcolor{black}{#1}}

\newcommand{\insteadof}[1]{\ignorespaces}

\title{\LARGE \bf
3DOF Pedestrian Trajectory Prediction Learned from Long-Term Autonomous Mobile Robot Deployment Data}

\author{Li Sun$^{1}$ and Zhi Yan$^{2}$ and Sergi Molina Mellado$^{1}$ and Marc Hanheide$^{1}$ and Tom Duckett$^{1}$
  \thanks{$^{1}$Lincoln Centre for Autonomous Systems (L-CAS), University of Lincoln, UK
    {\tt\small \{lsun, smolinamellado, mhanheide, tduckett\}@lincoln.ac.uk}}%
  \thanks{$^{2}$Laboratoire Electronique Informatique Image (Le2i), University of Technology of Belfort-Montb\'eliard (UTBM), France
    {\tt\small zhi.yan@utbm.fr}}%
}

\begin{document}

\maketitle
\thispagestyle{empty}
\pagestyle{empty}

\begin{abstract}

This paper presents a novel 3DOF pedestrian trajectory prediction approach for autonomous mobile service robots.
While most previously reported methods are based on learning of 2D positions in monocular camera images, our approach uses range-finder sensors to learn and predict 3DOF pose trajectories (i.e. 2D position plus 1D rotation within the world coordinate system).
Our approach, T-Pose-LSTM (Temporal 3DOF-Pose Long-Short-Term Memory), is trained using long-term data from real-world robot deployments and aims to learn context-dependent (environment- and time-specific) human activities. Our approach incorporates long-term temporal information (i.e.~date and time) with short-term pose observations as input. \kevinupdate{A sequence-to-sequence LSTM encoder-decoder is trained, which encodes observations into LSTM and then decodes as predictions.} For deployment, it can perform on-the-fly prediction in real-time. 
Instead of using manually annotated data, we rely on a robust human detection, tracking and SLAM system, providing us with examples in a global coordinate system. 
We validate the approach using more than 15K pedestrian trajectories recorded in a care home environment over a period of three months.
The experiment shows that the proposed T-Pose-LSTM model advances the state-of-the-art 2D-based method for human trajectory prediction in long-term mobile robot deployments.
\end{abstract}

\section{INTRODUCTION}
\label{sec:introduction}

Pedestrian trajectory prediction is still an open problem, especially for real-world deployments of autonomous mobile robots.
Most of the existing work focuses on the modeling of social interactions between people or human groups in large indoor public areas \cite{helbing1995social,UCY,ETH,IGP,yamaguchi2011you,linder2014multi,social-lstm,context-lstm}.
However, these approaches typically do not consider the spatial and temporal context of human activities, which can help trajectory prediction, especially over longer durations (e.g.~more than 5 seconds).
For instance, in a large hospital, pedestrians are likely to queue up in a reception area during specific periods in the daytime or at an emergency desk at night-time, and are likely to walk through corridors and stand near coffee machines.
In order to capture such contextual cues, long-term sensory data from actual robotic deployments can be used to learn predictive models.

Conventional approaches for pedestrian trajectory prediction are based on the learning of 2D trajectories from manually annotated data~\cite{UCY,ETH,IGP,yamaguchi2011you,social-lstm,context-lstm,SDD}.
Data-driven methods such as Social-LSTM \cite{social-lstm} achieve the state-of-the-art performance in 2D trajectory prediction.
The emerging 3D LiDAR devices are able to provide long-range and wide-angle laser scans, and are very accurate and not affected by lighting conditions.
With the continuing reduction in hardware prices, 3D LiDAR is becoming a popular choice for pedestrian detection and tracking for mobile robot applications.

\begin{figure}[t]
\centering
\includegraphics[width=0.5\textwidth]{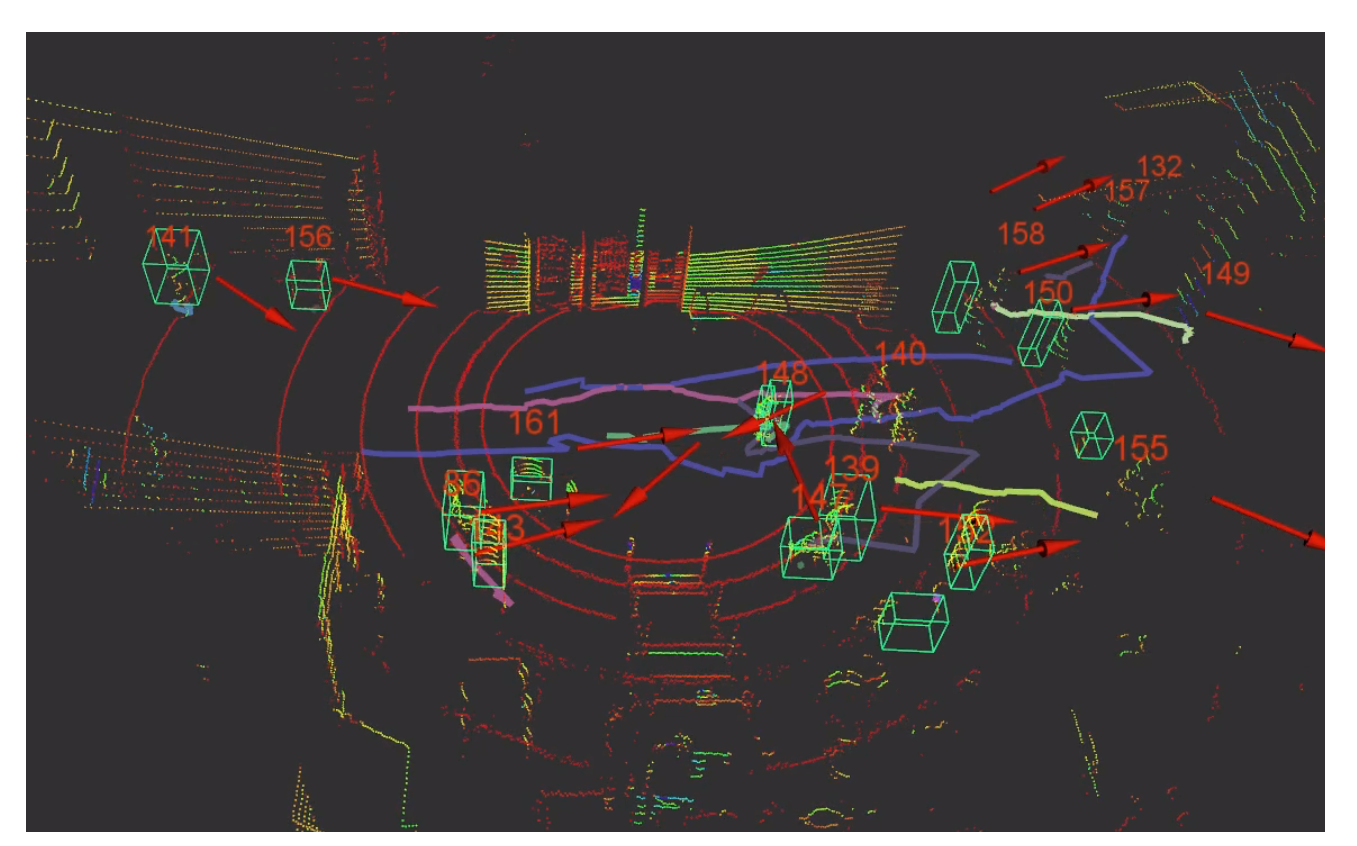}
\caption{A screen-shot of our 3DOF pedestrian trajectory prediction in a 3D LiDAR scan. The detected people are enclosed in green bounding boxes with a unique ID.
The colored lines represent the observed people trajectories.
The red arrow indicates the predicted poses for the next $1.2s$.}
\label{fig:pose_example}
\end{figure}

Most of the existing works on human trajectory prediction have the following limitations.
Firstly, the existing datasets were recorded with monocular cameras, which are generally used to predict pedestrian positions in image frames rather than in real-world coordinates.
Though in these datasets~\cite{UCY,ETH,SDD} the x-y coordinates are converted to the camera's coordinate system (in meters) using the intrinsic camera parameters, the depth is missing.
Secondly, the existing work does not consider the spatial and temporal context of the human activities in the environment.

In this research, we use range-finder sensors for human trajectory observation and prediction. Our model is trained using two datasets collected by mobile service robots (one using 2D laser and RGB-D sensors, and the other 3D laser/LiDAR), including a 3-month autonomous deployment in a care home~\cite{Hawes2016}.
The contributions of this paper are as follows.
We propose a novel approach for learning to predict 3DOF pedestrian trajectories (including both 2D position and orientation) from the real-world robot data, using a new LSTM-based architecture to incorporate spatial and temporal context information from the deployment environment.
We also publish our data, comprising a novel 3D pedestrian trajectory dataset based on 3D LIDAR data.
The dataset and demo are available online at: https://lcas.lincoln.ac.uk/wp/3dof-pedestrian-trajectory-dataset/.

The remainder of this paper is organized as follows:
Section~\ref{sec:related_work} gives an overview of the related literature;
Section~\ref{sec:methodology} describes our approach based on the LSTM model;
Section~\ref{sec:experiments} presents the experimental results on two real-world datasets;
and the paper is concluded with contributions and suggestions for future research.

\section{RELATED WORK}
\label{sec:related_work}
In this section, we first review the state-of-the-art approaches for pedestrian trajectory prediction and activity forecasting. Then we introduce the latest achievements on long-term autonomy of mobile robotic systems.

\subsection{Pedestrian Trajectory \& Activity Prediction}
Predicting pedestrian trajectories has a long history in computer vision~\cite{UCY,ETH,IGP,yamaguchi2011you,social-lstm,context-lstm,SDD}.
Pioneering work focuses on representing the social context with hand-crafted features~\cite{UCY,ETH,IGP,yamaguchi2011you}, while more recent approaches have tried to learn from context information relating to the position of other pedestrians or landmarks in the environment~\cite{social-lstm,context-lstm}.
As the trajectory prediction can be formulated as a regression problem, Gaussian Process (GP)-based methods~\cite{IGP,sparse-GP} are very popular in pedestrian prediction.
The limitation of GP-like parametric models is that the computation increases exponentially with the number of training examples.
Although sparse GP methods~\cite{sparse-GP} can be much more efficient for large-scale datasets, this problem is unlikely be remedied.

LSTM-based data-driven approaches~\cite{social-lstm,context-lstm} have achieved the state-of-the-art performance, especially with large-scale datasets~\cite{SDD}. 
Later, an end-to-end trajectory prediction method was proposed by \cite{varshneya2017human}, which combines the CNN (Convolutional Neural Network) for local image context understanding and LSTM for consecutive position prediction. 

Furthermore, researchers are working on understanding higher-level human behaviors from pedestrian trajectories in station surveillance videos, including work on travel time estimation~\cite{yi2015pedestrian}, recognition of stationary crowd groups~\cite{yi2016pedestrian}, and destination forecasting~\cite{alahi2014socially}.
Pioneering research has also considered the representation of dynamic maps from long-term datasets, 
including the FreMEn approach~\cite{fremen} for signal analysis, which uses the Fourier Transform to identify underlying periodic processes in the environment and hence make predictions of human activities based on the frequency spectrum of the observations.

\subsection{Person Detection and Tracking in Autonomous Mobile Robot Systems}
Nowadays, mobile robots are equipped with various sensors such as RGB/RGB-D/stereo cameras and 2D/3D LiDARs, alone or in combination.
An important use of these sensors is to detect and track objects, including pedestrians. 
For example, \cite{jafari14icra} used a template and the depth information of the RGB-D camera to identify human upper bodies (i.e. shoulders and head), and the classical Multiple Hypotheses Tracking (MHT) algorithm for pedestrian tracking.
\cite{arras07icra} extracted 14 features for legs detection and tracking in 2D LiDAR scans, including the number of beams, circularity, radius, mean curvature, mean speed, and more.
Both approaches used an offline trained model, while our previous work~\cite{yz17iros} learned a human model online in 3D LiDAR scans with the help of an EKF (Extended Kalman Filter)-based tracking system.
Regarding combined use of different sensors, \cite{linder16icra} introduced a people tracking system for mobile robots in very crowded and dynamic environments.
Their system was evaluated with a robot equipped with two RGB-D cameras, a stereo camera and two 2D LiDARs.
Although the above work has been able to track pedestrians in world coordinates ($xy$-plane), 
such approaches have limited applicability for trajectory prediction beyond 1 second.

Over the past 20 years of development of Simultaneous Localization and Mapping (SLAM), 2D laser-based SLAM \cite{gmapping} and visual-SLAM \cite{mur2015orb,endres20143} have largely become off-the-shelf technologies employed in many robotic systems. 
With the advances in robustness and adaptability of robotic systems in real-world environments, long-term autonomy has become an important research topic, starting from museum robots interacting with visitors~\cite{Thrun1999MINERVA:Robot}, through robots employed in office environments~\cite{Biswas2016TheResults,Marder-Eppstein2010TheEnvironment}, to robots running for months in care homes~\cite{Hawes2016}. Such interactive robots provide an excellent opportunity to learn from the long-term observations of people in their vicinity and gather -- and subsequently exploit -- such experience for their own behaviour generation and planning. Data sets gathered by the long-term autonomous robotic system of the STRANDS project~\cite{Hawes2016,Hanheide2017} are also used to evaluate the approach proposed in this paper (see Sec.~\ref{sec:datasets}).

\subsection{Discussion}
After investigation of the state-of-the-art methods for pedestrian trajectory prediction, we concluded that their limitations are twofold.
Firstly, the predominantly reported methods are based on monocular cameras, see e.g.\ the \ UCY~\cite{UCY}, ETH~\cite{ETH} and SDD~\cite{SDD} datasets.
The drawback of monocular cameras is that the trajectories are not in real-world coordinates (meters) when the camera is not perpendicular to the ground.
Also, the visual scope of monocular cameras is very narrow on mobile platforms. For mobile robot applications, alternative means of sensing need to be found.
Secondly, the existing works on pedestrian trajectory prediction proposed various methods to parameterize the social context, while few of them take account of the overall spatial and temporal context. 


In order to learn environment- and time-specific activity patterns, long-term data from the target environment, i.e.\ covering several weeks or more, is required. Recent state-of-the-art methods for real-time detection, tracking and SLAM have pushed the limits of mobile service robots towards life-long autonomy. The emerging 3D LiDAR sensors have the potential to be the sensor of choice for pedestrian activity prediction by mobile robots. In this paper, our contribution is to incorporate spatial and temporal context information into human trajectory prediction, by training our model using actual data from such long-term deployments.

\begin{figure*}[t]
\centering
\includegraphics[width=1.0\textwidth]{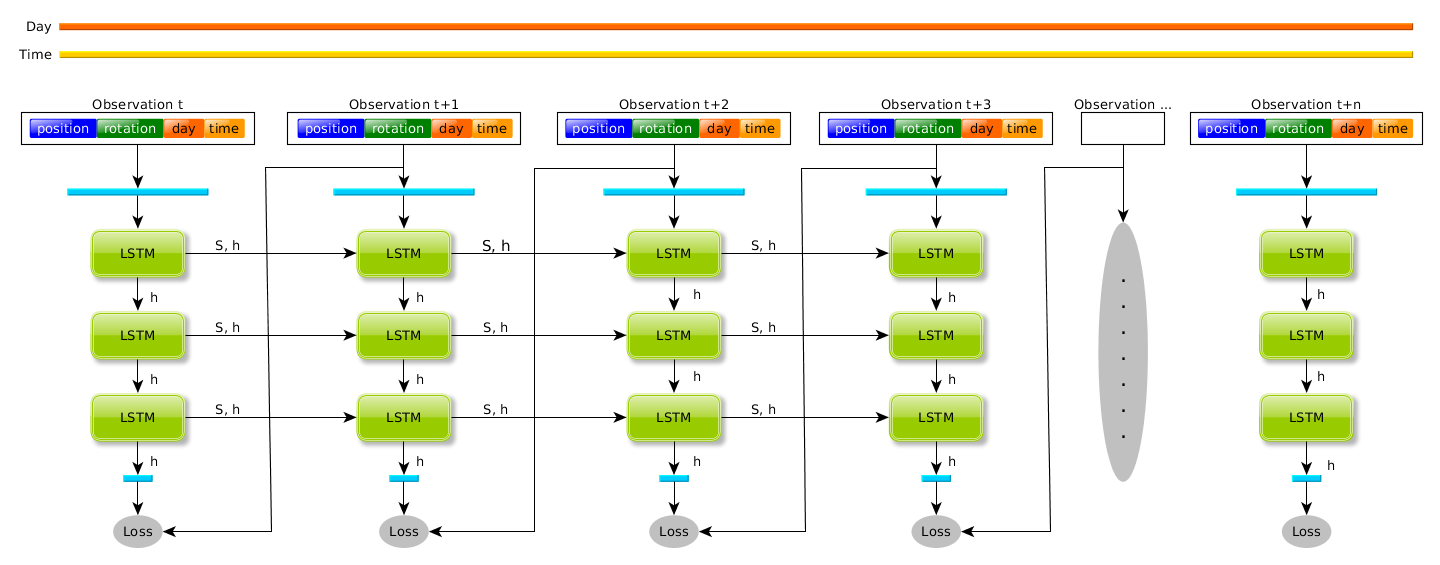}
\caption{The architecture of our Temporal Pose-LSTM network. A shared-triple-layer LSTM is trained in a sequence-to-sequence encoder-decoder form.  }
\label{fig:seq2seq}
\end{figure*}

\section{METHODOLOGY}
\label{sec:methodology}

\subsection{Problem Formulation}
Our trajectory prediction model is based on LSTM~\cite{lstm}.
We train an LSTM network on consecutive observations and predict the observation in the future time stamp.
Practically speaking, in order to enhance the supervision, we train the LSTM as an encoder-decoder between two sequences of observations with time difference $\Delta t$.
Given a sequence of observations $O:\{o_t, \dots, o_{t+n+1}\}$ with interval of $\Delta t$, the LSTM is modeled as a sequence-to-sequence encoder-decoder between $O_{\{t, t+n\}}:\{o_t, \dots, o_{t+n}\}$ and $O_{t+1, t+n+1}:\{o_{t+1}, \dots, o_{t+n+1}\}$.
The observation $o_t$ at time $t$ is a 6DOF pedestrian pose $\{x, y, z, q_x, q_y, q_z, q_w\}$ (here, $q_{.}$ are quaternions), which we simplify in our approach to 3DOF $\{x, y, q_z, q_w\}$ with the assumption that pedestrians have constant height (z-axis) in indoor environments and only $yaw$ axis rotation is applicable.
Finally, we concatenate the pose with additional time information, i.e.\ calendar day and hour-minute-second $\{t_{day}, t_{hms}\}$, and obtain the final observation $\{x, y, q_z, q_w, t_{day}, t_{hms}\}$.
It is worth noting that, instead of using camera frame data, we transform the coordinates to the world frame, and as a consequence, the two-dimensional position feature can represent the \kevinupdate{3D positions} of pedestrians in real-world coordinates.

\subsection{Network Architecture}
The architecture of the proposed LSTM for 3DOF Trajectory prediction is shown in Fig.~\ref{fig:seq2seq}. As shown in the figure, the 3DOF pose data are concatenated with time information as the observation. Before input to the LSTM, for each pedestrian trajectory, we encode the observations into a 128-dimensional embedding using a fully-connected layer and rectified linear unit (ReLU) activation function:

\begin{equation}
\small
(S_{t+1}, h_{t+1}) = LSTM\big(\phi(o_i; W_e), S_{t}, h_{t}, g_i, g_f, g_o; W_{\{i,f,o,s\}} \big),
\end{equation}

\noindent where  $W_e$ are the parameters in the embedding layer and $\phi$ is the non-linear function of the linear embedding with a ReLU activation. $S_{t}, h_{t}$ and $S_{t+1}, h_{t+1}$ are the LSTM's state and output variables of time $t$ and $t+1$. $g_i, g_f, g_o$ refers to the $input$, $forget$ and $output$ gate, respectively. $W_{\{i,f,o,s\}}$ are the parameters of LSTM. In our approach, the social-pooling reported in \cite{social-lstm} is not used as there were no substantial improvement in our testing scenarios, where there were few obvious social behaviors. 


Our LSTM model is a triple-layer sequence-to-sequence model with output size of 128. Using a multi-layer LSTM can enlarge the receptive field of the short-memory and thereby preserve the short-memory against vanishing too quickly. In our approach, we used a shared LSTM cell for all three layers in order to achieve a longer-term short memory without increasing the model's parameters. The LSTM is trained as an encoder-decoder from the pose at the current time $t$ to the pose of the next observation at $t+\Delta t$. Hence for each observation in the training sequence, the next observation is connected with the loss of LSTM as the prediction ground truth.

In the real-world data, the pedestrian trajectories are of dynamic length because of the limited sensor scope. In order to fully leverage all the trajectory data, we train our LSTM with dynamic sequence lengths. Practically speaking, we obtain a batch of training sequences of dynamic lengths, we use a binary activation mask to memorize the position $(row_i, col_i)$ of the observations and 
obtain the batch loss as:
\begin{equation}
loss_{batch} = \mathcal{\bf 1}_{i}[row_i, col_i] \odot [loss_i],
\end{equation}
\noindent where $\odot$ is an element-wise multiplication operation.

\subsection{3DOF Pose Loss}
We incorporate the position loss and the rotation loss into the loss function. More specifically, similar to previous works with 2D data \cite{social-lstm}, we use the Gaussian Probabilistic-Density-Function ($PDF$) as the likelihood function for position prediction. For the rotation loss, we tried both $L2$ loss and cosine distance loss. In this paper, $L2$ loss is used as it achieves the best performance. Moreover, a $L2$ regularization term is applied on all weights to eliminate over-fitting. Overall, our pose trajectory prediction loss function is:
\begin{equation}
\begin{split}
loss = \sum_i^N \sum_j^{n} -log \big ( PDF((x_{gt}, y_{gt})^{i,j},\mathcal{N}^{i,j}(\mu, \Sigma))  \big )\\
+ \Vert r_p^{i,j}-r_{gt}^{i,j}\Vert_2 + \lambda \Vert W \Vert_2, ~~~~~~~~~
\end{split}
\end{equation}
\noindent where $n$ is the length of the observation sequences, and $N$ is the number of training sequences. $(x_{gt}, y_{gt})^{i,j}$ is the ground truth x-y position and $r_p^{i,j}$=$(q_p^z, q_p^w)^{i,j}$ and $r_{gt}^{i,j}$=$(q_{gt}^z, q_{gt}^w)^{i,j}$ refer to prediction quaternions and ground truth quaternions, respectively. $W$ refers to all the trainable weights in our neural network and $\lambda$ is the weight of the regularization term (a fixed value of 0.005 is used in our implementation). A bi-variant Gaussian distribution is used in the $PDF((x_{gt}, y_{gt})^{i,j},\mathcal{N}^{i,j}(\mu, \Sigma))$ function:
\begin{equation}
\begin{split}
PDF=
\frac{exp \big (-\frac{1}{2}((x_{gt}, y_{gt})^{i,j}-\mu) \Sigma^{-1} ((x_{gt}, y_{gt})^{i,j}-\mu)^T \big )} {((2 \pi)^2|\Sigma|)^{-1/2}}
\end{split}
\end{equation}
\begin{equation}
\small
\mu = ( \mu_x, \mu_y )^{i,j}
, 
~ \Sigma = \Big( 
\begin{tabular}{cc} 
  $\sigma_x^2$ & $\rho \sigma_x \sigma_y$ \\
  $\rho \sigma_x \sigma_y$ & $\sigma_y^2$
\end{tabular}
\Big)^{i,j},
\end{equation}
\noindent where $\mu_x, \mu_y$ are two mean variables, $\sigma_x, \sigma_y$ are the two standard deviations and $\rho$ is the correlation between them. Within this loss function, our neural network decoder therefore has 7 outputs corresponding to $(\mu_x, \mu_y, \sigma_x, \sigma_y, \rho, q_p^z, q_p^w)^{i,j}$.

\subsection{Life-long Deployment}
Given a prediction $(\mu_x, \mu_y, \sigma_x, \sigma_y, \rho, q_p^z, q_p^w)^t$ from LSTM at time $t$, the forward prediction position at time $t+1$ can be estimated by sampling within the \kevinupdate{predicted} bi-variant Gaussian distribution:
\begin{equation}
x_{t+1}, y_{t+1} = \frac{1}{N_{s}} \sum_k^{N_{s}} (x_s, y_s)^k \sim \mathcal{N}_p(\mu, \Sigma),
\end{equation}
\noindent \kevinupdate{where $\mathcal{N}_p$ is obtained from $(\mu_x, \mu_y, \sigma_x, \sigma_y, \rho)^t$  }and $N_s$ is the number of samples. The rotation (quaternion) of $t+1$ can be directly obtained from the LSTM output $(0, 0, q_p^z, q_p^w)^t$ with normalization. For a long-term consecutive prediction, we use the new predicted pose as the input of the latest observation and predict the poses iteratively. 

In our implementation, we proposed an on-the-fly LSTM for real-time life-long pedestrian trajectories prediction. Our system is able to be deployed in real time (10~Hz)\footnote{In our video demo, we used a 2.5~Hz prediction rate, which is the standard time interval used widely in similar problems.}. For each iteration, all pedestrians in the robot's visual range are scanned, detected and tracked. The pedestrians' identities and poses are published with a certain frequency and subscribed by Pose-LSTM on-the-fly. Each pedestrian has its own LSTM states. We initialize the states of newly appearing pedestrians and free the states of disappeared pedestrians. For all the pedestrians at the same time stamp, we extract the pose features and update their states using a shared LSTM model. 

\section{EXPERIMENTS}

\label{sec:experiments}
In this section, we first introduce two novel 3D pedestrian activity datasets, the STRANDS pedestrian dataset and L-CAS pedestrian dataset. We further provide the details of our evaluation protocol and implementation. The experimental result on the STRANDS and L-CAS datasets are presented in Section \ref{sec:exp1} and Section \ref{sec:exp2}.

\subsection{Two Novel 3D Pedestrian Activity Datasets}\label{sec:datasets}

\begin{figure*}[t]
\centering
\includegraphics[width=0.98\textwidth]{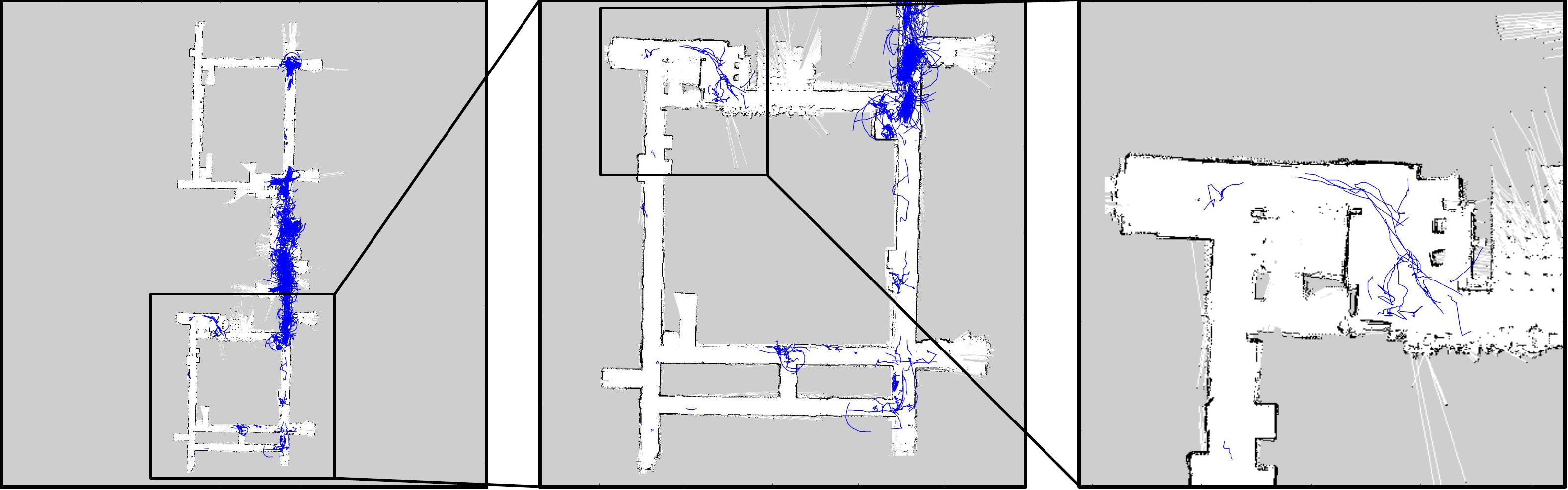}
\caption{Part of the trajectory examples in the STRANDS dataset are shown. The three sub-figures are zoomed in iteratively.}
\label{fig:strands_data}
\end{figure*}

\begin{figure*}[t]
\centering
\includegraphics[width=\textwidth]{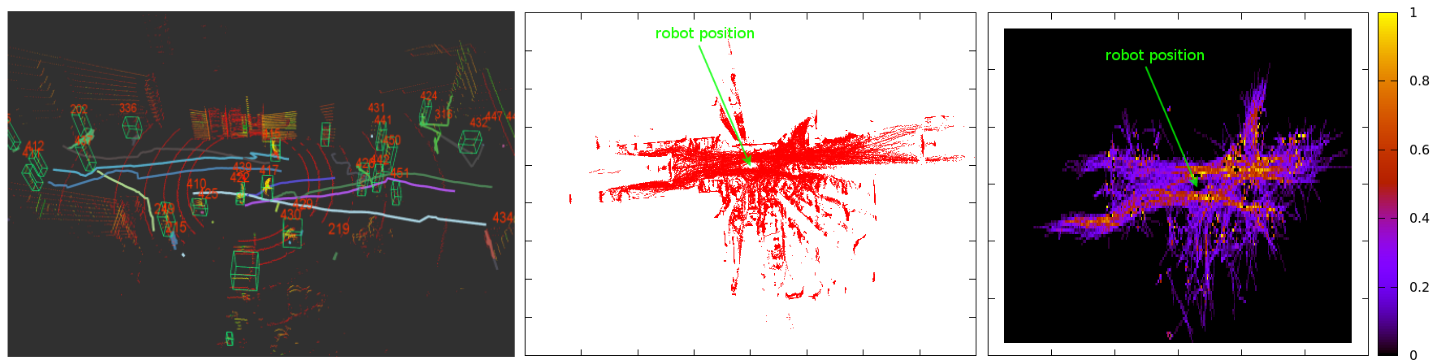}
\caption{Trajectory examples in the L-CAS dataset including: extracted pedestrian trajectories (left), detected point clusters (middle), and trajectories heatmap (right). In the heatmap, warmer colors indicate higher frequencies of pedestrian occupancy. The map is normalized between 0 and 1.}
\label{fig:flobot_data}
\end{figure*}

We evaluated our approach with two recent datasets which have 3DOF pose annotations, i.e. the STRANDS~\cite{Hawes2016} person trajectory dataset (see Fig.~\ref{fig:strands_data}), and the L-CAS people Dataset~\cite{yz17iros} (see Fig.~\ref{fig:flobot_data}), rather than with the conventional 2D image datasets~\cite{UCY,ETH,SDD}.

The STRANDS dataset was collected during a long-term robot deployment in a care facility in Vienna, Austria, spanning all weeks from 28/11/2016 until 3/4/2017 ($19$ weeks). During this time the robot traveled a total of more than 87 km along the corridors of the ground floor of the care home, as sketched in Fig.~\ref{fig:strands_data} and served in 6866 different tasks, interacting with humans. 
The robot employed is based on a SCITOS G5 platform, equipped with a 2D Sick Laser and two ASUS Xtion Pro cameras (one on the head for person detection and tracking, and one on the chest for obstacle avoidance); for details refer to~\cite{Hawes2016}.  For pedestrian detection, a kinect-based upper-body detector, and 2D laser-based leg detector  are integrated in an unscented Kalman-Filter tracking framework, as described in~\cite{Dondrup2015a}. In total, 17609 trajectories (comprising timestamp, global $x$ and $y$ coordinates on the floor plane, as well as the estimated angle of the direction of travel) of persons encountered in this environment are recorded and the average length of each trajectory is 22.6 seconds.


The L-CAS dataset was collected by a Velodyne VLP-16 3D LiDAR, mounted at a height of 0.8~m from the floor on the top of a Pioneer 3-AT robot, in one of the main buildings (a large indoor public space, including a canteen, a coffee shop and the resting area) of Lincoln University, UK.
This dataset captures new research challenges for indoor service robots including human groups, children, people with trolleys, etc.
Similar to the STRANDS dataset, the data were recorded in the sensor reference frame, and all human detections and tracks were then transformed to the world frame.
We conducted our experiments on the first 19 minutes of data, in which 935 pedestrian trajectories were extracted.
A comparison with the above-mentioned prior datasets can be seen in Table~\ref{tab:datasets}.

\begin{table*}[t]
\centering
\caption{Comparison of the existing datasets for pedestrian trajectory prediction}
\label{tab:datasets}
\begin{tabular}{|l|l|l|l|l|l|l|}
\hline
Dataset & Duration & \#Tracks & Ave. Len. & Sensors & Views & Annotation\\
\hline
ETH~\cite{ETH} & $<$ 1h & 390 & 6.7s & RGB camera (building view) & camera-view & manual \\
\hline
UCY~\cite{UCY} & $<$ 1h & 434 & 16.5s & RGB camera (building view) & camera-view & manual \\
\hline
SDD~\cite{SDD} & $\approx$ 8.6h & 11216 & - & RGB camera (bird view) & camera-view & auto \\
\hline
STRANDS~\cite{Hawes2016} & $\approx$ 3192h & 17609 & 22.6s & RGB-D camera, 2D LiDAR & global-view & auto  \\
\hline
L-CAS~\cite{yz17iros} & $<$ 1h & 935 & 13.5s & 3D LiDAR & global-view & auto \\
\hline
\end{tabular}
\end{table*}

\begin{table}[thpb]
\centering
\caption{Comparison of the pedestrian trajectory approaches on the STRANDS dataset. The $ADE \vert AEDE$ are shown.}
\label{tab:exp_result}
\begin{tabular}{|l|l|l|l|}
\hline
\backslashbox{Test}{Method} & Social-LSTM\cite{social-lstm} & Pose-LSTM & T-Pose-LSTM \\
\hline
obs. 5s pred. 1s    &   0.48m $\vert$ nan  &    0.43m $\vert$ 3.8$^\circ$     &     \bf{0.43m} $\vert$ 3.8$^\circ$  \\
\hline
obs. 5s pred. 2s    &  0.68m  $\vert$ nan  &    0.62m $\vert$ 4.9$^\circ$     &    \bf 0.53m $\vert$  4.6$^\circ$ \\
\hline
obs. 5s pred. 3s  &   0.89m  $\vert$ nan &     0.79m $\vert$ \textbf{5.6$^\circ$}    &    \textbf{0.76m} $\vert$  5.8$^\circ$ \\
\hline
obs. 5s pred. 4s  &   1.13m  $\vert$ nan &     0.93m $\vert$  \textbf{6.2$^\circ$}   &     \textbf{0.87m} $\vert$  6.7$^\circ$ \\
\hline
obs. 5s pred. 5s &   1.28m $\vert$ nan  &    1.03m $\vert$ \textbf{6.7$^\circ$}    &     \textbf{0.90m} $\vert$   6.9$^\circ$\\
\hline
obs. 5s pred. 6s &   1.41m $\vert$ nan  &    1.42m $\vert$  8.9$^\circ$    &     \bf 1.09m $\vert$  7.6$^\circ$ \\
\hline
obs. 5s pred. 7s &   1.62m $\vert$ nan  &   1.35m $\vert$ 8.0$^\circ$     &     \bf 1.31m $\vert$   8.6$^\circ$\\
\hline
obs. 5s pred. 8s &   1.75m  $\vert$ nan &   1.32m  $\vert$ \textbf{7.2$^\circ$}     &    \textbf{1.25m} $\vert$   8.8$^\circ$\\
\hline
obs. 5s pred. 9s &   1.95m $\vert$ nan  &   1.47m $\vert$ 8.1$^\circ$     &    \textbf{1.38m $\vert$  9.3$^\circ$} \\
\hline
\end{tabular}
\end{table}

\begin{table}[thpb]
\centering
\small
\caption{Comparison of the pedestrian trajectory approaches on the L-CAS dataset. The $ADE \vert AEDE$ are shown.}
\label{tab:exp2}
\begin{tabular}{|p{2.3cm}|p{2.3cm}|p{1.7cm}|}
\hline
Dataset Methods & Social-LSTM\cite{social-lstm} & Pose-LSTM \\
\hline
L-CAS & 1.19m $\vert$ nan & 0.95m $\vert$ 35$^\circ$  \\
\hline
\end{tabular}
\end{table}

\begin{figure}[t]
\centering
\includegraphics[width=0.5\textwidth]{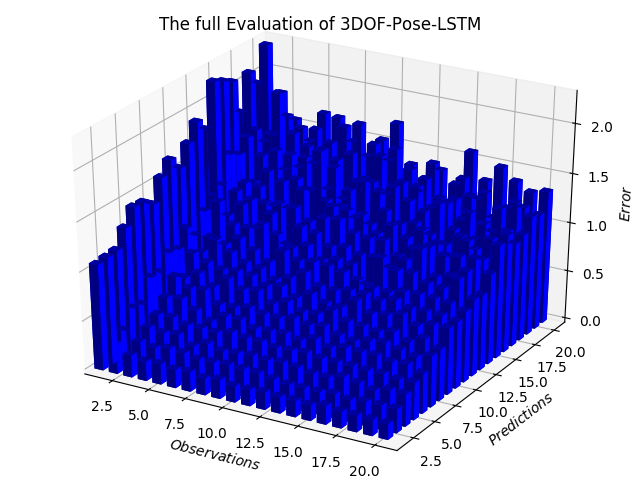}
\caption{A full evaluation of 3D-Pose-LSTM on the L-CAS dataset is shown. In this figure, the length of observations/prediction sequence varies from 1 to 20 (with the interval of 0.4s) and the errors are in meters. }
\label{fig:exp2_mat}
\end{figure}
\subsection{Evaluation Protocol}
Following the previous works \cite{ETH,IGP,social-lstm,context-lstm,SDD}, the Average Displacement Error ($ADE$) is used to measure the error of predicted positions, which is computed using mean square error over all predicted position and ground truth positions as: 
\begin{equation}
\small
ADE = \frac{1}{N*n} \sum_i^{N} \sum_j^{n} \sqrt{(x_p^{i,j}-x_{gt}^{i,j})^2 + (y_p^{i,j}-y_{gt}^{i,j})^2}
\end{equation}

As our approach predicts the 3DOF pose rather than 2D position, the Average Eulerian angle Difference error ($AEDE$) is used to measure the rotation loss. We first convert the prediction and the ground truth quaternion to Eulerian angles, and calculate the absolute error of the $yaw$ angle as:
\begin{equation}
\small
\begin{split}
AEDE = \frac{1}{N*n} \sum_i^{N} \sum_j^{n} min( \vert euler_z(r_{p}^{i,j}) - euler_z(r_{gt}^{i,j}) \vert, \\
2\pi - \vert euler_z(r_{p}^{i,j}) - euler_z(r_{gt}^{i,j}) \vert ),~~~~~~~~~~~~~~~~
\end{split}
\end{equation}
\noindent where the $euler_z$ is the function converting quaternions to Eulerian angle (i.e. $yaw$). In this paper, the measurement of $ADE$ is in meters and $AEDE$ is in degree ($\pi$/180).

For comparison, as there are no previous works for 3DOF pose trajectory prediction, we implement the state-of-the-art 2D trajectory prediction method Social-LSTM~\cite{social-lstm} as a baseline method. We use Social-LSTM as the baseline as it outperforms most of the conventional methods, e.g.~\cite{UCY,ETH,IGP,yamaguchi2011you,SDD,grewal2011kalman}. As both Social-LSTM and our approach are LSTM-based methods, this comparison can indicate the difference between 2D position prediction (Social-LSTM), the proposed 3DOF pose prediction (Pose-LSTM) and Time-included 3DOF Pose-prediction (T-Pose-LSTM).

\subsection{Implementation Details}
In our LSTM implementation (including Social-LSTM), we use the 128-dimensional feature embedding and the hidden state dimension of LSTM is 128. For Social-LSTM, we follow its original configuration on social-pooling: spatial pooling size is 32 and pooling window size is 8$\times$ 8. For the training, we use \kevinupdate{a mini-batch of 128 with} RMS-prop optimizer~\cite{bengiormsprop} for training. There are a few differences between our Social-LSTM implementation and the original LSTM reported in \cite{social-lstm}: Firstly, we use a sequence-to-sequence model to enhance supervision. Secondly, a dynamic sequence length is used in training. Thirdly, we do not use the synthetic data generated by Social-Force \cite{helbing1995social} for pre-training. Our implementation is based on the TensorFlow library\footnote{https://www.tensorflow.org/}. More details of training are shown in Section \ref{sec:exp1} and \ref{sec:exp2}.

\subsection{Experiments on STRANDS dataset}\label{sec:exp1}
In this experiment, we split the whole dataset into 2/3 for training and 1/3 for testing randomly on the time axis, which means that the training sequences and testing sequences are fully split. As a result, we get 11743 frames for training and 5866 frames for testing. The training comprises two steps: we first train LSTM with a fixed sequence length of 20 for 100 epochs with an initial learning rate of 0.005 with exponential decay of 0.98. Then we finetune the LSTM with a dynamic length of [8, 20] for another 100 epochs with a learning rate 0.003 and exponential decay of 0.98.

More specifically, the settings for the inputs of the LSTM models are as follows: Social-LSTM: 2 dimensional position i.e. $x$, $y$,  Pose-LSTM: 4 dimensional pose i.e. $x$, $y$, $r_z$, $r_w$; T-Pose-LSTM: 4 dimensional pose and 2 dimensional time representation, i.e.\ $x$, $y$, $r_z$, $r_w$, $date$, $time$. We normalize each input dimension to $\mathcal{N}(0, 1)$ as pre-processing. In this experiment, a single-layer LSTM is used rather than triple-layers due to the large observation interval (1s). For our proposed approach, both the Average Displacement Error ($ADE$) and Average Eulerian angle Difference error ($AEDE$) are evaluated. 
  
In order to evaluate the effectiveness of our proposed approach, we take 5 observations and predict the following 1 to 9 seconds. The observations are made with a frequency of 1HZ. As shown in Table \ref{tab:exp_result}, on the STRANDS dataset our proposed T-Pose-LSTM achieves the lowest $ADE$ among the three LSTM models. For a short-term prediction (less than 5 seconds), the performance of three methods are very close, while after 5 seconds, T-Pose-LSTM experienced a substantial advantage for the longer-term prediction. This verified our hypothesis, that the time information learned from long-term data can improve activity prediction. Without time information, Pose-LSTM still performs better than the baseline (approximately 10\%-20\%) on $ADE$, so we conclude that the encoded orientation (pose) information has the potential to enhance the position prediction. For the orientation prediction, T-Pose-LSTM and Pose-LSTM produce similar $AEDE$, and we did not give the $AEDE$ of Social-LSTM as it is proposed for 2D position prediction. 

\subsection{Experiments on L-CAS Pedestrian Trajectory Dataset}\label{sec:exp2}

In this experiment, the L-CAS 3D pedestrian dataset is used. We acquire the 3DOF pose trajectories from the tracker. Following the settings of previous research~\cite{UCY, ETH,social-lstm,context-lstm,SDD}, we sample the frames at 2.5HZ (time interval of 0.4s). We take 8 observations (3.2 seconds) and predict the following 12 observations (4.8 seconds). Similar to the experiments on the STRANDS dataset, we split the whole dataset into 2/3 of the trajectories for training and 1/3 for testing. In this experiment, the triple-layer shared LSTM architecture is used as the interval of observations is very short. 
The training has two steps: we first train Pose-LSTM with a fixed sequence length of 30 for 100 epochs, then finetune with a dynamic sequence length [10, 20] for another 100 epochs. As this dataset only covers around 20 minutes, long-term training data is not available, so we only compared our proposed Pose-LSTM with the baseline method (Social-LSTM \cite{social-lstm}) without including date and time information.

The result on the L-CAS dataset is shown in Table \ref{tab:exp2}, where our proposed Pose-LSTM experienced a good improvement (0.95m $AVE$) compared to the baseline (1.19m $AVE$). The $AEDE$ of Pose-LSTM is 35 degrees, which is much higher than for the STRANDS dataset. This is because there are more static humans in the L-CAS dataset and the \kevinupdate{poses annotations} of static people estimated by the Bayesian tracker are not accurate. We further fully verified the proposed model on different combinations of observation length and prediction length, with the results shown in Fig. \ref{fig:exp2_mat}. We can observe that the predictions are of poor accuracy when the observation length is 1 as LSTM cannot give reliable predictions with only the initial state input. For the remaining combinations, the $ADE$ error increases gradually with increasing prediction length and decreases gradually with more observations.

\section{CONCLUSION}\label{sec:conclusion}
In this paper, we presented a novel neural network model Pose-LSTM for predicting the 3DOF pose trajectories rather than 2D positions of humans. Compared to 2D models, e.g.\ Social-SLTM, our Pose-LSTM is able to predict more information, i.e.\ both position and orientation. Predicting higher dimensional data is a more challenging task for data-driven methods, while our approach utilized the correlation between different sources of input and is able to predict more information without losing accuracy. For a longer-term prediction, our proposed T-Pose-LSTM learned from long-term robot deployment data incorporates the short-term observation and long-term spatio-temporal context, which makes longer-term prediction more accurate.

For future work, we will extend our novel 3D LiDAR trajectory dataset (L-CAS dataset) to a period of several weeks across several buildings. Furthermore, we will investigate the possibility of training our Pose-LSTM in life-long mode for open-ended learning in dynamic environments. 





\section*{ACKNOWLEDGMENT}
We thank NVIDIA Corporation for generously donating a high-power GPU on which this work was performed. This project has received funding from the European Union's Horizon 2020 research and innovation programme under grant agreement No 732737 (ILIAD).

\bibliographystyle{IEEEtran}
{
\bibliography{refs,marcs_mendeley}}


\end{document}